\documentclass[letterpaper,10pt,conference]{ieeeconf}
\IEEEoverridecommandlockouts   
\usepackage[english]{babel}
\usepackage[colorinlistoftodos]{todonotes}
\usepackage{amsmath}
\usepackage{lipsum} 
\usepackage[T1]{fontenc}
\usepackage{bm} 
\usepackage[linesnumbered,ruled,vlined]{algorithm2e}
\usepackage{amssymb}
\usepackage[utf8]{inputenc}
\usepackage{graphicx}
\usepackage{graphics}
\usepackage{subcaption}
\usepackage{siunitx}
\usepackage{multirow}
\usepackage{float}
\usepackage{caption}
\usepackage{gensymb}
\usepackage{epstopdf}
\usepackage{enumerate}
\usepackage{multicol,multirow}
\usepackage{booktabs}
\usepackage{cite}
\usepackage{balance}
\usepackage{tikz,xcolor}
\usepackage{academicons}
\usepackage{multirow}
\usepackage{adjustbox}
\usepackage{siunitx}
\usepackage{bigints}
\usepackage{dsfont}

\makeatletter
\let\NAT@parse\undefined
\makeatother
\usepackage[hidelinks]{hyperref}
\DeclareCaptionLabelSeparator{periodspace}{.\quad}
\usepackage{mathtools}
\DeclarePairedDelimiterX{\infdivx}[2]{(}{)}{%
  #1\;\delimsize\|\;#2%
}
\newcommand{\infdiv}{D\infdivx}
\DeclareMathOperator{\E}{\mathbb{E}}

\usepackage{svg}
\definecolor{orcidlogocol}{HTML}{A6CE39}
\graphicspath{{figures/}} 
\definecolor{lime}{HTML}{A6CE39}
\DeclareRobustCommand{\orcidicon}{%
	\begin{tikzpicture}
	\draw[lime, fill=lime] (0,0) 
	circle [radius=0.16] 
	node[white] {{\fontfamily{qag}\selectfont \tiny ID}};
	\draw[white, fill=white] (-0.0625,0.095) 
	circle [radius=0.007];
	\end{tikzpicture}
	\hspace{-2mm}
}

\foreach \x in {A, ..., Z}{
	\expandafter\xdef\csname orcid\x\endcsname{\noexpand\href{https://orcid.org/\csname orcidauthor\x\endcsname}{\noexpand\orcidicon}}
}

\hypersetup{
    pdfauthor   = {Daniel Layeghi },%
    pdftitle    = {ICRA_PAPER},%
    pdfsubject  = {Robotics},%
    pdfkeywords = {Optimization, Approximate inference, Monte Carlo Sampling, Path Integral, Differential Dynamic Programming}%
}

\begin{document}
\bstctlcite{IEEEexample:BSTcontrol}
	\title{\LARGE\bf Optimal Control via Combined Inference and Numerical Optimization}
	\author{Daniel Layeghi, Steve Tonneau and Michael Mistry
		\thanks{All authors are with the School of Informatics, University of Edinburgh. Email: d.layeghi@sms.ed.ac.uk}
		\thanks{}
	}
\maketitle

\begin{abstract}
Derivative based optimization methods are efficient at solving optimal control problems near local optima. However, their ability to converge halts when derivative information vanishes. The inference approach to optimal control does not have strict requirements on the objective landscape. However, sampling, the primary tool for solving such problems, tends to be much slower in computation time. We propose a new method that combines second order methods with inference. We utilise the Kullback Leibler (KL) control framework to formulate an inference problem that computes the optimal controls from an adaptive distribution approximating the solution of the second order method. Our method allows for combining simple convex and non convex cost functions. This simplifies the process of cost function design and leverages the strengths of both inference and second order optimization. We compare our method to Model Predictive Path Integral (MPPI) and iterative Linear Quadratic Gaussian controller (iLQG), outperforming both in sample efficiency and quality on manipulation and obstacle avoidance tasks.

\end{abstract}
	
	
\IEEEpeerreviewmaketitle
	
\section{INTRODUCTION} \label{sec:Introduction}
The goal of minimising an objective is a general one that applies to many areas. In the simplest cases the age-old solution is to solve for the minimisation analytically. In many interesting cases though the analytical method becomes intractable and we have to resort to computational methods. Methods for computing the minimisation generally fall into two categories: numerical and sampling based optimization. Both methods have advantages when applied to the correct domain. Numerical methods are better suited to domains where the objective of concern is smooth and derivative information is available. On the other hand, if the objective function is highly discontinuous and multi-modal, sampling is a better choice due to its inherent exploitative capabilities and looser requirements on the objective. Interesting optimization problems however are not only restricted to one of the mentioned domains. As a result, it may seem desirable to combine both methods to work in tandem. However, the abstraction that combines both methods and formulation of a consensus between both methods is nontrivial. This paper aims to naturally combine the use of numerical methods to refine when dense derivative information is available and sample to explore when derivatives vanish. 

Here we focus on the optimal control setting and its application to robotic, interesting settings where the governing dynamics are discontinuous and the exact structure of the objective function is unknown. For example, locomotion, is a task where the higher level objective is to transfer from point A to B. The exact objective function that defines the intricate behaviour to achieve the goal is unknown and the derivative of this cost with respect to the span of controls is usually sparse. As a result making and breaking contacts to solve the problem creates discontinuous optimization landscapes. One can imagine the same principles in dexterous manipulation problems or conversely problems where avoiding contacts is key, for example, obstacle avoidance.
\section{Related Work}
One approach to remove the problem of sparse derivatives is to relax the optimization problem. \cite{mordatch2012discovery} show impressive results on locomotion and manipulation tasks. The key insight is the relaxation of the contact constraints. Resulting in dynamics where contact forces exist at distance. A drawback of this approach is the problem of local minima. Once the optimization landscape is sufficiently smoothed there may exist many local optima where finding the feasible is hard. In addition physically unrealistic solutions are also a problem. To mediate the problem of local optimum \cite{toussaint2020describing} introduced an adaption with additional logical decision variables over action modes such as sliding and flying. This allows for selective smoothing of dynamics reducing the total number of local optima. However, the choice modes create a limiting range of dynamics. In addition, the problem remains deterministic leading to extremely intricate behaviours that are not robust.

To tackle the sparsity problem \cite{posa2014direct} introduces linear complementary constraints to the optimization problem this allows the optimizer to decide whether bodies are in contact. This formulation underpins many formulations around locomotion. \cite{pardo2017hybrid, winkler2018gait,cebe2020online} use the same framework with additional task specific parametrization. The formulation in this approach is specific to contact dynamics and additionally the optimization generated is large and slow.

\cite{aceituno2017simultaneous,kuindersma2016optimization} proposed a separate approach to reducing this problem of sparsity. They decompose the problem of planning and control and use integer variables and Mixed Integer Programming to select the dynamic modes of systems, for example, in or out of contact. Subsequently a trajectory optimization problem is then solved to follow the planned points. The major drawback of this method is the loose relationship between the two solvers. As a result, significant work has been done to introduce dynamic feasibility into the planning segment\cite{tonneau2020sl1m}. Such feasibility constraints can be highly task-dependent and difficult to formulate.

Apart from numerical methods. \cite{kappen2012optimal,todorov2007linearly} proposed an entirely different approach inspired by Kalman's duality \cite{stengel1986stochastic} that treats the optimal control problem as an inference. In this formalisation, the dynamics are treated as graphical models and the optimal control distribution is solved through approximate inference by using a variant of Monte Carlo methods. \cite{toussaint2009robot} extended this approach by introducing Gaussian approximations to solve nonlinear problems. \cite{theodorou2010generalized} proposed a path integral (PI) formulation of this approach. \cite{rajamaki2016sampled} proposed a PI inspired approach modified by insights from differential dynamic programming (DDP) \cite{jacobson1968new}. The inference approach results in desirable outcomes such as clear formulations for exploration and the relaxation of the requirement on the objective such as continuity and smoothness, therefore, simpler cost function design for achieving goals. However, the major drawback of such methods is one of sampling, i.e. inferring the unknown distribution is computationally costly. 

We aim to combine the strength of both inference and second order trajectory optimization. We aim to formulate a natural combination that can efficiently sample when no derivative information is available and follow optimized trajectories when derivatives arise.

\section{The Optimal Control Setting} \label{sec:Methodology}
Let us consider the generic discrete optimal control setting. An agent at state $\mathbf{x} \in \mathbf{X}$ is required to choose an action or control $\mathbf{u} \in \mathbf{U}$ such that the resulting sequence of states and actions $\{(\mathbf{x}_0, \mathbf{u}_0), (\mathbf{x}_1, \mathbf{u}_1) ... (\mathbf{x}_n)\}$ minimises the summation associated to the total cost of each state and action pair $J(\mathbf{X}, \mathbf{U})$. This problem can be solved by using the \textit{principle of optimality}. That is the optimal solution is the combined optimal solutions to the subproblems of the original problem. This principle is formalised in the Bellman equation \cite{bellman1966dynamic} which underpins most of the optimal control theory.  
\begin{equation}
J(\mathbf{x}, t) = \min_{\mathbf{u}} {\left[l(\mathbf{x}, \mathbf{u}, t) + J\left({\text{f}(\mathbf{x}, \mathbf{u}, t)}\right) \right]}  \nonumber
\end{equation}

$J(\mathbf{x}, t)$ represents the value function, $l(\mathbf{x}, \mathbf{u})$ the running cost and $f(\mathbf{x}, \mathbf{u})$ the deterministic system dynamics all evaluated at $\mathbf{x}$ and $\mathbf{u}$. Solving this equation in its vanilla form suffers from curse dimensionality; the computation grows exponentially with the number of states. In the next sections, we focus on the numerical and inference based solution to the problem.

\subsection{Numerical Optimization of The Optimal Control}\label{sec: Num opt}
The most established numerical solution to the Bellman solves the problem in its tangent space using a second order method. This formulation is known as Differential Dynamic Programming (DDP)\cite{jacobson1968new}. Rewriting the Bellman equation as a function of state and control and using $Q(\mathbf{x_t}, \mathbf{u_t})$ to represent the summation of running cost with the value function
\begin{equation}\label{eq: bellman}
J(\mathbf{x}, t) = \min_{\mathbf{u}} [Q(\mathbf{x}_t, \mathbf{u}_t)]  \nonumber
\end{equation}
Making a second order Taylor approximation leads to
\begin{equation}
\begin{aligned}
Q(\delta{\mathbf{x}}, \delta{\mathbf{u}}) &\approx \frac{1}{2}{\begin{bmatrix}1 \\ \delta{\mathbf{x}} \\ \delta{\mathbf{u}} \end{bmatrix}}^{\intercal}\begin{bmatrix} 0 & Q_{\mathbf{x}}^{\intercal} & Q_{\mathbf{u}}^{\intercal} \\
Q_{\mathbf{x}} & Q_{\mathbf{ux}} & Q_{\mathbf{ux}} \\ 
Q_\mathbf{u} & Q_{\mathbf{ux}} & Q_{\mathbf{uu}} \end{bmatrix} \begin{bmatrix} 1 \\\delta{\mathbf{x}} \\ \delta{\mathbf{u}} \end{bmatrix} \nonumber
\end{aligned}
\end{equation}
Here the subscripts represents derivatives. Minimising the approximation with respect to the perturbed control $\delta{\mathbf{u}}$ and solving for the optimal control update
\begin{equation}
\begin{aligned}
&Q_\mathbf{u} + Q_{\mathbf{ux}}\delta{\mathbf{x}} + (\hat{\mathbf{u}} - \mathbf{u})Q_{\mathbf{uu}} = 0 \\
&\hat{\mathbf{u}} = \mathbf{u} - \frac{Q_\mathbf{u}}{Q_{\mathbf{uu}}} -\frac{Q_{\mathbf{ux}}}{Q_{\mathbf{uu}}}\delta{\mathbf{x}}
\nonumber
\end{aligned}
\end{equation}
The above is identical to Newton's method where the $Q_\mathbf{u} + Q_{\mathbf{ux}}\delta{\mathbf{x}}$ is the control gradient direction and $Q_{\mathbf{uu}}$ is the Hessian with respect to control used as curvature information. In DDP the update step is defined through a feed-forward gain $\textbf{k} = Q_\mathbf{u}{Q^{-1}_{\mathbf{uu}}}$ and a feedback gain $\textbf{K} = Q_{\mathbf{ux}}{Q^{-1}_{\mathbf{uu}}}$. \cite{tassa2012synthesis} is a comprehensive explanation of the Gauss-Newton version of DDP known as iLQG. In this case, the second order derivatives with respect to dynamics are ignored and additional regularization terms are introduced for the robustness of Hessian approximations. From here onwards we refer to iLQG when referring to the numerical solution to the optimal control.
\subsection{Approximate Inference of Optimal Control}\label{sec: approx inf}
The inference approach to the optimal control problem has been tackled under many formulations. The KL control approach is grounded in the Bellman equation therefore drawing comparisons with the numerical approach is simpler. As a result, our focus in this paper is the KL-control approach and its dualities with path integral control. One can view the KL control approach to be grounded in the stochastic version of the Bellman equation shown in section \ref{sec: Num opt}

\begin{center}
$\underset{\mathbf{u}}{\text{min}}\left[l(\mathbf{x}) + \infdiv{\mathbf{p}(\mathbf{x}'|\mathbf{x}, \mathbf{u})}{\mathbf{p}(\mathbf{x}'|\mathbf{x})}+\E_{\mathbf{x}^{'}\sim \mathbf{u}}\left[J(\mathbf{x}')\right]\right]$
\end{center}

Where $p(\mathbf{x}'|\mathbf{x}, \mathbf{u})$ and $p(\mathbf{x}'|\mathbf{x})$ represent the controlled and uncontrolled probabilities, $D$ represents a KL divergence operator. KL divergence measures the information difference between two distributions. $J$ is the typical value function. Under this formulation the probability for the optimally controlled dynamics becomes

\begin{center}
$\mathbf{p}^*(\mathbf{x}'|\mathbf{x}, \mathbf{u}) = \mathbf{p}(\mathbf{x}'|\mathbf{x})\text{exp}\left(-J(\mathbf{x}')\right)\eta^{-1}$
\end{center}

Therefore the optimal transition probability $\mathbf{p}^*(\mathbf{x}'|\mathbf{x}, \mathbf{u})$ becomes the transition probability of the passive dynamics adjusted by the exponential values of states and normalised by $\eta$.

KL control's central assumption is that the stochasticity of dynamics is entirely the function of noise over the control input. Meaning, in our case with normal input noise, an input $\mathbf{u}_t$ becomes $\mathbf{v}_t \sim \mathcal{N}(\mathbf{u}_t, \Sigma)$. As a result if the control input trajectory $V = (\mathbf{v}_0, \mathbf{v}_1, ..., \mathbf{v}_{T-1})$ and the initial state $\mathbf{x}_0$ are known the resulting states trajectory can be computed. This effect allows for representing the state distributions in terms of the control distributions. We can represent the state distributions by computing the likelihood, where $X = (\mathbf{x}_i, ..., \mathbf{x}_{tk}, ..., \mathbf{x}_{T})$

\begin{equation}
\begin{aligned}
\mathbf{p}^*(X) &= \prod_{k=i}^{N}\mathbf{p}(\mathbf{x}_{t_{k+1}}|\mathbf{x}_{t_k})\text{exp}\left(-J(\mathbf{x}_{t_{k+1}})\right)\eta^{-1}\\
&=\mathbf{p}(X)\text{exp}\left(-J(X)\right)\eta^{-1}
\nonumber
\end{aligned}
\end{equation}

Using the same state-control mappings defined in \cite{williams2017information}, optimal state distribution $\mathbf{p}^*(X)$ can be represented in terms of optimal control trajectory $\mathbf{q}^*(V)$ under passive and candidate control distributions; $\mathbf{p}(V) := \mathbf{v}_t \sim \mathcal{N}(0, \mathbf{\Sigma}),~\mathbf{q}(V) := \mathbf{v}_t \sim \mathcal{N}(\mathbf{u}_t, \mathbf{\Sigma})$

\begin{equation}
\mathbf{q}^*(V) =\mathbf{p}(V)\text{exp}\left(-J(V)\right)\eta^{-1}
\nonumber
\end{equation}

\cite{theodorou2012relative,williams2017information,theodorou2015nonlinear} show comprehensive derivation and a Model Predictive Control (MPC) formulation that allows to importance sample from this distribution. Our final algorithmic implementation uses the overall MPC structure of this method shown in \cite{williams2017information}.

\section{Combining Sampling With Second Order Method} \label{sec:Methodology}
To combine second order optimization and approximate inference, we start by reformulating the Bellman equation and introduce a secondary divergence term with the solution of the Bellman equation in the tangent space. That is the optimal trajectory obtained by DDP/iLQG algorithm. Modifying the KL objective we can write
\begin{equation}
\begin{aligned}\label{eq: combined Bellman}
&\underset{\mathbf{u}}{\text{min}}[l(\mathbf{x}) + (1-k) \infdiv{\mathbb{Q}}{\mathbb{P}} + k \infdiv{\mathbb{Q}}{\mathbb{C}} + \E_{\mathbb{Q}}[J(\mathbf{x}')]]\\
&\infdiv{\mathbb{Q}}{\mathbb{P}} = \E_{\mathbb{Q}}\left[\log\left[\frac{\mathbf{p}(\mathbf{x}'|\mathbf{x}, \mathbf{u})}{\mathbf{p}(\mathbf{x}'|\mathbf{x})}\right]\right]\\
&\infdiv{\mathbb{Q}}{\mathbb{C}} = \E_{\mathbb{Q}}\left[\log\left[\frac{\mathbf{p}(\mathbf{x}'|\mathbf{x}, \mathbf{u})}{\mathbf{p}(\mathbf{x}'|\mathbf{x}, \mathbf{u}_{\text{iLQG}})}\right]\right]
\end{aligned}
\end{equation}
\noindent where $\mathbb{P}$ is the distribution of the passive dynamics for which the input $v$ has a mean of 0 with variance $\mathbf{\Sigma}$. $\mathbb{C}$ represents the distribution of the controlled dynamics under iLQG where the input has a mean of $\mathbf{u}_{\text{iLQG}}$ with variance $\mathbf{\Sigma}_{\text{iLQG}}$. $\mathbb{Q}$ represents our control distribution which has a mean of $u$ and variance $\Sigma$ under the input $v$. $k \in [0, 1]$ is the importance given to minimising each KL. 

Combining expectations and rearranging leads to the optimal control trajectory (complete derivation in the Appendix \ref{appendix})
\begin{equation}
\begin{aligned}\label{eq: combined opt dist}
\mathbf{q}^*(V) = \mathbf{p}(V)^{1-k}\mathbf{c}(V)^{k}\text{exp}\big(-\frac{1}{\lambda}J(V)\big)\eta^{-1}
\end{aligned}
\end{equation}
Where $\lambda$ is a hyperparameter discounting trajectory costs. Equation \ref{eq: combined opt dist} is very similar to the optimal transition described in section \ref{sec: approx inf}. However in this case the optimal trajectory is a factor of the joint distributions $\mathbb{P}$ and $\mathbb{C}$ scaled by the importance $k$ and adjusted by the exponentiated negative cost of trajectories $\text{exp}(S(-V))$. In similar fashion to \cite{williams2017information} we use equation \ref{eq: combined opt dist} to importance sample from the optimal distribution $\mathbb{Q^*}$
\begin{equation}\label{eq: importance sampling}
\mathbf{u}^* = \bigintsss\mathbf{q}(V)\underbrace{\frac{\mathbf{q}^*(V)}{\mathbf{p}(V)^{1-k}\mathbf{c}(V)^{k}}\frac{\mathbf{p}(V)^{1-k}\mathbf{c}(V)^{k}}{\mathbf{q}(V)}}_{w(V)} v_t \mathrm{d} V
\end{equation}

\begin{figure*}[]
  \includegraphics[width=\textwidth,height=4cm]{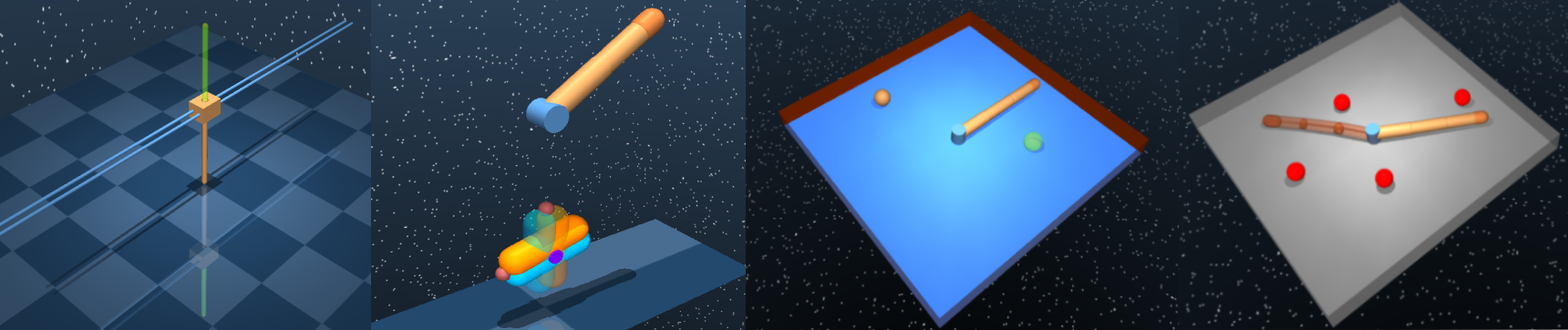}
  \caption{Left to right: Cartpole scenario: The objective is to swing the pole to upwards position. Finger-Spinner: The objective is for the fully actuated 2 link manipulator to rotate the object to the desired location marked as transparent. Push Object: The goal is for the 3-link manipulator to move the orange sphere to the desired location (green). Obstacle Avoidance: The goal is for the 3 link manipulator to move to the desired position (red) without making contact with spherical obstacles. Results video at \url{https://youtu.be/dZgRJ0CTpLw}}
  \label{fig: envs}
\end{figure*}

\begin{table*}[]
\centering
\caption{Total trajectory cost and success rate for each method per environment. The costs are computed across length of trajectories until completion of tasks. Column Time represents the total computation time until task completion in seconds. For number of samples and hyperparameters refer to each task refer to section B.}
\begin{adjustbox}{max width=\textwidth}
\begin{tabular}{ccccccccccccc}
\hline
\multirow{2}{*}{Environment \%} &\multicolumn{3}{c}{Ours}&\multicolumn{3}{c}{MPPI}&\multicolumn{3}{c}{Naive Combination}&\multicolumn{3}{c}{iLQG}\\
\cline{2-13}
&Cost& \% Success&Time&Cost& \%  Success&Time&Cost& \%  Success&Time&Cost& \%  Success&Time\\
\hline
Cartpole & \num{6.69e+03}               & 100& 15.05 &\num{8.20e+03}& 100&10.03&\num{2.07e4}   & 100 &33.13& \num[math-rm=\mathbf]{2.5e3}& 100&1.84\\
Finger-Spinner     & \num[math-rm=\mathbf]{1.45e8}& 100&7.19 &\num{1.97e8}  & 100&7.68&\num{3.52e8}  & 100 &10.56&N/A                         & 0&N/A\\
Push Object     & \num[math-rm=\mathbf]{2.12e11}& 100& 15.87&\num{4.92e11}  & 100 &73.93&N/A  & 0&N/A& N/A                         & 0&N/A\\
Obstacle Avoidance & \num[math-rm=\mathbf]{6.50e5}& 100 &12.95&\num{8.79e5}  & 25 &11.376&N/A & 0&N/A& N/A & 0&N/A\\
\hline
\end{tabular}
\end{adjustbox}
\label{tabel: Result Comp}
\end{table*}

Computing the importance sampling weights $w(V)$ with respect to the distributions (see Appendix \ref{appendix}) for a trajectory in length of time T. 
\begin{equation}
\begin{aligned}\label{eq: sample weights}
w(V) =& \frac{1}{\eta}\exp\bigg(-\frac{1}{2}\frac{1}{\lambda}S(V)\\
&+\frac{1}{2}\sum_{t=0}^{T}\Big(\mathbf{u}^T_{t}\mathbf{\Sigma}_{t}^{-1}\mathbf{u}_{t} + 2\mathbf{v}^T_{t}\mathbf{\Sigma}_{t}^{-1}\mathbf{u}_{t}\\
&+ k\left({(\mathbf{v}_{t}-\mathbf{u}_{\text{iLQG}~t})}^T\mathbf{\Sigma}_{\text{iLQG}_t}^{-1}(\mathbf{v}_{t}-\mathbf{u}_{\text{iLQG}_t})\right)\\
&-k({\mathbf{v}_t}^T\mathbf{\Sigma}_{t}^{-1}\mathbf{v}_t)\Big)\bigg)
\end{aligned}
\end{equation}

From equation \ref{eq: sample weights} $\infdiv{\mathbb{Q}}{\mathbb{C}}$ translates into a cost term within $w(V)$ that penalises the deviation of the sampled control $\mathbf{v}_t$ from the solution obtained by iLQG, $\mathbf{u}_{\text{iLQG}_t}$ where the cost gain is the covariance of the iLQG solution $\mathbf{\Sigma}_{\text{iLQG}_t}$. Additionally, the parameter $k$ dictates the importance of this cost and if set to 0 the weights become the same as the KL control case. $\mathbf{\Sigma}_{\text{iLQG}_t}$ is a proxy measure of our confidence in the solution provided by the second order method. 
To compute the variance of $\mathbb{C}$, the distribution that represents the solution of the second order method, we know that the trajectory $\mathbf{u}_{\text{iLQG}~t}$ minimises the quadratic approximation of the instance of the \textit{Q} function $\delta Q(\mathbf{x}_t, \mathbf{u}_{t})$ at $\mathbf{u}_{\text{iLQG}_t}$. As a result, the mean and the variance for the normal distribution $\mathbb{C}$ can be defined by maximizing the log-likelihood:
\begin{equation}
\begin{aligned}\label{eq: ML}
\ln \exp (-\delta Q(\mathbf{x}_t, \mathbf{u}_t))
\end{aligned}
\end{equation}
maximising the log-likelihood leads to the same problem shown in section \ref{sec: Num opt}. Therefore, $\frac{\delta}{\delta \mathbf{u}} \ln \exp (-\delta Q_t(\mathbf{x}_t, \mathbf{u}_t))$ is at a maximum if  $\mathbf{u}_{t} = \mathbf{u}_{\text{iLQG}_t}$. We can obtain the variance by computing the second derivative, $\frac{\delta^2}{\delta \mathbf{u}^2} \ln \exp (-\delta Q_t(\mathbf{x}_t, \mathbf{u}_t)) = Q_{\mathbf{uu}_t}^{-1} = \mathbf{\Sigma}_{\text{iLQG}_t}$. As a result input trajectories sampled form $\mathbb{C}$ take the form $v_t \sim \mathcal{N}(\mathbf{u}_{\text{iLQG}_t}, Q_{\mathbf{uu}_t}^{-1})$. In our final formulation we scale this variance approximation with a hyperparameter $\beta$.

Algorithm \ref{alg: InTO} shows the final structure of our proposed method. KL control method here represents the algorithm in \cite{williams2017information} with updated importance weights \ref{eq: sample weights}.

\begin{algorithm}[]
\SetAlgoLined
$K$: \text{Number of samples} and $T$: \text{Number of time-steps}\\
$q_{\text{iLQG}_r}(\textbf{x}_t)$ and $q_{r}(\textbf{x}_t)$: \text{Running costs}\\
$q_{\text{iLQG}_t}(\textbf{x}_T)$ and $q_{t}(\textbf{x}_T)$: \text{Terminal costs}\\
$\textbf{u}_{0:T-1}$ and $\textbf{u}_{\text{iLQG}_{0:T-1}}$: Initial trajectories\\

$\lambda,~k,~\beta,~\mathbf{\Sigma}$: Hyper parameters for eq. \ref{eq: sample weights}\\
\While{not done}
{
    $\mathbf{x}_0$ = EstimateState()\\
    $[\textbf{u}_{\text{iLQG}_{0:T-1}}, ~Q_{\mathbf{uu}_{0:T-1}}]$ = iLQG($\textbf{x}_0$)\\
    $\mathbf{\Sigma}_{\text{iLQG}_{0:T-1}} = \beta ~Q^{-1}_{\mathbf{uu}_{0:T-1}}$\\
    $\textbf{u}_{0:T-1}$ = KLControl($\textbf{u}_{\text{iLQG}_{0:T-1}},~\mathbf{\Sigma}_{\text{iLQG}_{0:T-1}},~\textbf{x}_0$)\\
    apply $\textbf{u}_0$\\
    $\textbf{u}_{0:T-1} = \textbf{u}_{1:T-1}$\\
    $\textbf{u}_{\text{iLQG}_{0:T-1}} =\textbf{u}_{0:T-1}$
}
\caption{Inference with Optimization (Ours)}
\label{alg: InTO}
\end{algorithm}

\section{Experimental Results and Discussion} \label{Experimental_Validation}

The results obtained were generated from environments that were designed to test the validity and benefits of our method. Therefore, key attributes of these environments shown in figure \ref{fig: envs} are
\begin{itemize}
    \item Discontinuous dynamics.
    \item Tasks where making or avoiding contact is necessary for completion.
    \item Tasks that allows for designing simple non convex cost functions that induce sparsity in derivative information.
\end{itemize}

In this section, we evaluate whether our algorithm can combine trajectory optimization and inference in a manner that allows for following the optimized trajectory when derivative information is available and resorting to unbiased inference when derivatives vanish. We compare our method to 3 other algorithms: MPPI, iLQG and the naive combination of both where the output of iLQG is used as an initial trajectory for MPPI and vice versa. The results obtained are obtained across 20 random seeds. For each task, the lengths of the trajectories between solvers are the same. Table \ref{tabel: Result Comp} is a summary of these results.  

\subsection{Implementation details} \label{Experimental_Validation}
The implementation was developed in C++. The environments were simulated using Mujoco 2.0 \cite{todorov2012mujoco}. We have tuned the contact dynamics in Mujoco to be realistic, additionally, our iLQG implementation uses derivative information at contacts. Some of the environments were modified versions of models developed by \cite{tassa2018deepmind}. The results were obtained on a PC with an Intel Core i9 CPU running Ubuntu 18.04.

\subsection{Cartpole task} 
The Cartpole is a task we use to evaluate and compare the performance of our method in its simplest case. In this task the state $\mathbf{x} \in \mathbb{R}^4$ consists of the cart and pole position and velocities. The desired state is $\mathbf{x}_\text{goal} = (0, 0, 0, 0)$. The running cost for this task is $(\textbf{x}_{\text{error}~t}) \mathbf{Q}_t (\textbf{x}_{\text{error}~t})$ and a terminal cost of $(\textbf{x}_{\text{error}~T}) \mathbf{Q}_T (\textbf{x}_{\text{error}~T})$ where $\mathbf{Q}_t = \text{diag}(\num{1e3}, \num{5e2},0,0)$ and $\mathbf{Q}_T = \text{diag}(\num{1e5}, \num{5e4}, \num{5e2},  \num{5e2})$ with horizon $T = 0.75s$ and number of samples $K = 3$. The hyperparameters $\beta$, $\lambda$ and $k$ were set to $\num{1e-3}$, 0.1 and 1 respectively.

The results show that iLQG outperforms all other methods, beating ours by a factor of 2.7. This is not surprising as the cost function is not sparse and dynamics are continuous leading to a very smooth optimization landscape where any derivative based method will do well. Our method however outperforms MPPI by a factor of 1.2. This is achieved with a very low number of samples and biasing the samples towards the iLQG trajectory by setting a low value of $\beta$.

\subsection{Finger-Spinner task}
This is a manipulation task where the states $\mathbf{x} \in \mathbb{R}^6$ are the joint positions of the manipulator and the passive spinner followed by their velocities. The desired goal state is $\mathbf{x}_\text{goal} = (0, 0, 0, 0, 0, 0)$. The cost function is quadratic with respect to the states where the gain terms act on the state of the spinner. $\mathbf{Q}_t = \text{diag}(0, 0, \num{1.5e4}, 0, 0, 50)$ and $\mathbf{Q}_T = \text{diag}(0, 0, \num{1.5e4}, 0, 0, 50)$. An additional binary scaling term $\mathds{1}(\textbf{x}_t)\textbf{x}_{\text{error}~t}$ is also added to the running cost,  where

\begin{equation}
  \mathds{1}(\textbf{x}_t)=\left\{
  \begin{array}{@{}ll@{}}
    0, & \text{if} ~ \text{contact between spinner and finger} \\
    1, & \text{otherwise.}
  \end{array}\right.
  \nonumber
\end{equation} 

This term tracks the global positions of bodies and encourages contact until low state errors. Since this term is non differentiable, it is only added to the inference cost in equation \ref{eq: sample weights}.
The choice of the hyperparameters are $[\beta, \lambda, k] = [0.25, 0.5, 1]$ and $T = 0.75$ and $K_{\text{MPPI}} = 30$ and $K_{\text{Ours}} = 10$

Our method outperforms all other methods, beating the second best (MPPI) by a factor of 1.3 with a third of the number of samples and faster compute cycle. The derivative based method (iLQG) entirely fails to solve this task. This is because the cost function is sparse and the dynamics are highly discontinuous. As a result, very few control inputs will create the contacts and the derivative direction required to solve the task. The naive combination is capable of solving this problem in effect by randomising the control input until contact is made. This results in a suboptimal trajectory, which is due to the very loose coupling between inference and numerical optimization. 

\subsection{Push Object task}\label{sec: push ob}
This is another manipulation task similar to the previous. In this task the object to be manipulated has higher degrees of freedom and the exploration domain for finding contact is much larger. The states $\mathbf{x} \in \mathbb{R}^{10}$ are the joint positions of the planar manipulator in addition to the global Cartesian position of the object that is manipulated followed by their derivatives. The desired state is $\mathbf{x}_\text{goal} = (0, 0 ,0, 0.25, -0.22, 0.022, 0, 0, 0, 0)$. The cost function is quadratic with respect to the states and only defined by a term on the state of the object where $\mathbf{Q}_t = \text{diag}(0, 0, 0, \num{1e5}, \num{1e5}, 0, 0, 0, \num{5e2}, \num{5e2})$ and $\mathbf{Q}_T = (0, 0, 0, \num{1e5}, \num{1e5}, 0, 0, 0, 0, 0)$. To encourage contact similar to the previous case, an additional binary term $\mathds{1}(\textbf{x}_t)\textbf{x}_{\text{error}~t}$ is added to the running cost in equation \ref{eq: sample weights}. The choice of the hyperparameters are $[\beta, \lambda, k] = [1, 0.25, 1]$ and $T = 0.75$, $K_{\text{MPPI}} = 80$ and $K_{\text{Ours}} = 50$ 

Our method performs better than all other methods. Beating the second best (MPPI) by a factor 2.3. Our result show that contact dynamics is the major contributing factor to this. Our method is capable of pushing the object much more intricately by staying close to the solution of the second order method. iLQG on its own is not capable of solving this task due to the same sparsity of derivatives and discontinuity reasons as the previous task. The naive combination also fails 25\% of the time and otherwise computes a trajectory that is an order of magnitude worse.

\subsection{Obstacle Avoidance task}\label{sec obs av}
This task is essentially the opposite of the manipulation task. For successful completion contact with environment has to be avoided. The states $\mathbf{x} \in \mathbb{R}^{6}$ are the joint positions and velocities. The desired state is $\mathbf{x}_\text{goal} = (9.72, 0, 0, 0, 0, 0)$. The cost function is quadratic with respect to the states, defined by a term on the state of the manipulator where $\mathbf{Q}_t = \text{diag}(100, 10, 10, 0, 0, 0)$ and $\mathbf{Q}_T = (100, 100, 100, 10, 1, 1)$. To discourage contact, a binary term penalising contacts $\mathds{1}(\textbf{x}_t)\textbf{x}_{\text{error}~t}$ is added to the running cost in equation \ref{eq: sample weights}. The choice of the hyperparameters are $[\beta, \lambda, k] = [20, 0.1, 1]$ and $T = 0.75$ , $K_{\text{MPPI}} = 100$ and $K_{\text{Ours}} = 50$ 

Similar to the previous case we outperform all other algorithms, on average beating MPPI by a factor of 1.4 when compared to its best case. We are able to achieve 100\% success rate in comparison to the 25\% success rate of MPPI with half the number of samples. By choosing lower values of $\lambda$ our algorithm performs complex maneuvers to escape local optima produced by the obstacles where as MPPI is not capable of generating such motions. iLQG and the naive combination are not able to avoid obstacles and are not capable of solving this task.

\section{Discussion}
The relationship between the covariance and the inverse Hessian $\beta Q_{\mathbf{uu}_t}^{-1} = \mathbf{\Sigma}_{\text{iLQG}_t}$ plays a central role in our method. We use this information to define a confidence measure. However, this measure simply reflects the rate of convergence and not the quality of the solution. This poses problems where derivative based optimization will pay high costs to obtain fast convergence.

The results show that our method is capable of finding derivative dense locations and following optimized trajectories. This notion is tightly coupled with the choice of hyperparameters, specifically $\lambda$ and $\beta$. This choice becomes apparent for environments with flat optimization landscapes such as the ones related to the tasks shown in section \ref{sec: push ob}. These cases require exploration through high values of $\lambda$ to allow for the sparse spread of the probability mass across trajectories. This however results in discounting the KL divergence cost which can be detrimental to high confidence optimized trajectories. Therefore choosing the hyperparameters in our method is not trivial and can result in unsuccessful trajectories.

The ability to introduce non differentiable costs to the inference part of our algorithm introduces a desirable formulation where we can remove the need to define complex convex costs for the derivative based part of our algorithm. For example, in section \ref{sec obs av} we remove the requirement for a distance cost to avoid obstacles. Another benefit of this formulation is the ability to encourage contacts with similar cost structures and removing requirements such as defining contact locations or introducing relaxations to guide the derivative based method towards contacts.

The results that we have obtained are for cases where the mixing term $k$ between the passive and the controlled distribution is set to 1. This is because $k$ induces a weighted joint distribution between the probabilities $\mathbb{P}$ and $\mathbb{C}$. This can become problematic when the distance between the two distributions is large. In that case, the joint can exist in a domain where no useful control inputs exist. An automatic adaption of the mixing term can prove useful when the control distribution is uninformative.

With regards to the computation time we show that we are able to outperform MPPI for our tasks. This is because of the sample efficiency of our algorithm through the combination with the second order method. This is especially apparent in cases where contact and reasoning about their dynamics is required. In such scenarios the second order method can provide more information around completion of the task. Our push object task shows this. We perform on similar wall clock times when compared to the naive implementation. This is expected as complexity of both algorithms are the same.
\section{Conclusion and Future Work} \label{Experimental_Validation}
We present a method that combines approximate inference with second order trajectory optimization. We formulate the approximate inference through a variant of  KL control that introduces a divergence term from the distribution of optimized trajectories. Through our derivation, we show that this distribution can be approximated by the solution of the trajectory optimization and its inverse Hessian. This natural combination removes the design of complex convex cost functions. We demonstrate this effect on our obstacle avoidance and manipulation tasks where we only use binary cost functions to encourage or discourage making contacts. We test our methods on environments with discontinuities and a mix of simple differentiable and non differentiable cost functions that nonetheless induce sparsity in derivative information. We compare our results to ones obtained from vanilla iLQG, iLQG with randomized warmstarts and MPPI. We show that in cases where derivative information is not immediately present, our method outperforms all others by exploring to find low cost trajectories and following the optimized trajectories when derivative information appears.
Some avenues can extend this work. The hyperparameters for temperature, scaling the Hessian and the mixing term between two distributions are central to efficient exploration and refinement. A systematic approach for an adaption of these parameters can bring efficiency gains. The KL control formulation of the approximate inference is based on the assumption of constant input noise. Introducing theoretically sound variance adaption strategies can provide better convergence especially when combined with derivative based optimization.

\appendix
\section{Appendix}
\subsection{KL control derivation}\label{appendix}

To obtain the final optimal distribution obtained in section \ref{sec:Methodology} we start by
\begin{equation}
\begin{aligned}
&\underset{\mathbf{u}}{\text{min}}[l(\mathbf{x}) + (1-k) \infdiv{\mathbb{Q}}{\mathbb{P}} + k \infdiv{\mathbb{Q}}{\mathbb{C}} + \E_{\mathbb{Q}}[J(\mathbf{x}')]]\\
&\infdiv{\mathbb{Q}}{\mathbb{P}} = \E_{\mathbb{Q}}\left[\log\left[\frac{\mathbf{p}(\mathbf{x}'|\mathbf{x}, \mathbf{u})}{\mathbf{p}(\mathbf{x}'|\mathbf{x})}\right]\right]\\
&\infdiv{\mathbb{Q}}{\mathbb{C}} = \E_{\mathbb{Q}}\left[\log\left[\frac{\mathbf{p}(\mathbf{x}'|\mathbf{x}, \mathbf{u})}{\mathbf{p}(\mathbf{x}'|\mathbf{x}, \mathbf{u}_{\text{iLQG}})}\right]\right]
\nonumber
\end{aligned}
\end{equation}
Combining expectations give (only minimisable terms of interest shown)
\begin{equation}
\begin{aligned}
&\underset{\mathbf{u}}{\text{min}}\left[ \E_{\mathbb{Q}}\left[\log\frac{\mathbf{p}(\mathbf{x}'|\mathbf{x}, \mathbf{u})}{\mathbf{p}(\mathbf{x}'|\mathbf{x})^{(1-k)}\mathbf{p}(\mathbf{x}'|\mathbf{x}, \mathbf{u}_{\text{iLQG}})^k\exp(-J(\mathbf{x}'))}\right]\right]\\
\nonumber
\end{aligned}
\end{equation}
The exponentiated negative values are not are not a transition probabilities therefore introducing a normalisation constant
\begin{equation}
\begin{aligned}
\phi(\mathbf{x}) = \sum \mathbf{p}(\mathbf{x}'|\mathbf{x})^{(1-k)}\mathbf{p}(\mathbf{x}'|\mathbf{x}, \mathbf{u}_{\text{iLQG}})^k \exp(-J(\mathbf{x}'))
\nonumber
\end{aligned}
\end{equation}
Adding the normalization constant to the original problem and assuming $z(\mathbf{x'}) = \exp(-V(\mathbf{x'}))$we can write
\begin{equation}
\begin{aligned}
&\underset{\mathbf{u}}{\text{min}}\left[\infdiv{\mathbf{p}(\mathbf{x}'|\mathbf{x}, \mathbf{u})}{\frac{\mathbf{p}(\mathbf{x}'|\mathbf{x})^{(1-k)}\mathbf{p}(\mathbf{x}'|\mathbf{x}, \mathbf{u}_{\text{iLQG}})^kz(\mathbf{x'})}{\phi(\mathbf{x})}}\right]\\
\nonumber
\end{aligned}
\end{equation}
The above KL is minimised when the two probabilities are equal:
\begin{equation}
\begin{aligned}
\mathbf{p}^*(\mathbf{x}'|\mathbf{x}, \mathbf{u}) = \frac{\mathbf{p}(\mathbf{x}'|\mathbf{x})^{(1-k)}\mathbf{p}(\mathbf{x}'|\mathbf{x}, \mathbf{u}_{\text{iLQG}})^kz(\mathbf{x}')}{\phi(\mathbf{x})}\\
\nonumber
\end{aligned}
\end{equation}

Moving to likelihoods

\begin{equation}
\begin{aligned}
\mathbf{p}^*(X) = \prod_{j=i}^{T}\frac{\mathbf{p}(\mathbf{x}_{t_{j+1}}|\mathbf{x}_t)^{1-k}\mathbf{p}(\mathbf{x}_{t_{j+1}}|\mathbf{x}_t, \mathbf{u}_{{\text{iLQG}}_t})^kz(\mathbf{x}_{t_{j+1}})}{\phi(\mathbf{x}_t)}\\
\nonumber
\end{aligned}
\end{equation}

Using  the  same  state-control  mappings  defined  in \cite{williams2017information} we write the above state distribution in terms of their density functions over a trajectory of length $T$ with control input $V$
\begin{equation}
\begin{aligned}
\mathbf{p}(V) = {}& Z^{-1}\exp\left(-\frac{1}{2}\sum_{t=0}^{T_1}\mathbf{v}_t^{T}\mathbf{\Sigma}^{-1}\mathbf{v}_t\right) \\
\mathbf{q}(V) = {}& Z^{-1}\exp\left(-\frac{1}{2}\sum_{t=0}^{T_1}(\mathbf{v}_t - \mathbf{u}_t)^T\mathbf{\Sigma}^{-1}(\mathbf{v}_t - \mathbf{u}_t)\right) \\
\mathbf{c}(V) = {}& Z^{-1}_{\text{iLQG}}\notag\\
&\exp\left(-\frac{1}{2}\sum_{t=0}^{T_1}(\mathbf{v}_t - \mathbf{u}_{\text{iLQG}_t})^T\mathbf{\Sigma}^{-1}_{\text{iLQG}_t}(\mathbf{v}_t - \mathbf{u}_{\text{iLQG}_t})\right)
\nonumber
\end{aligned}
\end{equation}
We represent optimal state distribution as a function of the above distributions. As a result the final optimal distribution becomes
\begin{equation}
\begin{aligned}
\mathbf{q}^*(V) &= \mathbf{p}(V)^{1-k}\mathbf{c}(V)^{k}\text{exp}\big(-\frac{1}{\lambda}S(V)\big)\eta^{-1}
\nonumber
\end{aligned}
\end{equation}
where $\eta$ is the normalisation constant over trajectories and $\lambda$ is the temperature an adjustable hyperparameter.

\bibliography{references}
\bibliographystyle{IEEEtran}
\balance

\end{document}